# Concept-Oriented Deep Learning with Large Language Models

Daniel T. Chang (张遵)

*IBM (Retired)* dtchang43@gmail.com

**Abstract:** Large Language Models (LLMs) have been successfully used in many natural-language tasks and applications including text generation and AI chatbots. They also are a promising new technology for concept-oriented deep learning (CODL). However, the prerequisite is that LLMs understand concepts and ensure conceptual consistency. We discuss these in this paper, as well as major uses of LLMs for CODL including concept extraction from text, concept graph extraction from text, and concept learning. Human knowledge consists of both symbolic (conceptual) knowledge and embodied (sensory) knowledge. Text-only LLMs, however, can represent only symbolic (conceptual) knowledge. Multimodal LLMs, on the other hand, are capable of representing the full range (conceptual and sensory) of human knowledge. We discuss conceptual understanding in visual-language LLMs, the most important multimodal LLMs, and major uses of them for CODL including concept extraction from image, concept graph extraction from image, and concept learning. While uses of LLMs for CODL are valuable standalone, they are particularly valuable as part of LLM applications such as AI chatbots.

## 1 Introduction

*Concept-oriented deep learning (CODL)* [1-2] is a machine learning approach that extends deep learning with *concept representations* and *conceptual understanding capability*. CODL is based on the idea that *concepts* are the foundation of human deep learning, understanding, and knowledge integration and transfer. *CODL systems* are composed of three main components: *concept graphs*, concept representations, and concept representation learning systems.

A *large language model (LLM)* [3-4] is a *deep learning model* with many parameters (typically billions of weights or more), trained on large quantities of *unlabeled text* using self-supervised learning or semi-supervised learning. LLMs have been successfully used in many natural-language tasks and applications including text generation and AI chatbots.

LLMs also are a promising new technology for CODL. However, the prerequisite is that LLMs understand concepts and ensure conceptual consistency. We discuss these in this paper, as well as major uses of LLMs for CODL including concept extraction from text, concept graph extraction from text, and concept learning.

One of the key challenges in developing LLMs is *concept understanding*. LLMs need to be able to understand the meaning of words and phrases in order to generate accurate and meaningful text. However, this can be difficult, as many words and phrases have multiple meanings. In particular, LLM's understanding of *abstract concepts* is significantly weaker than that of *concrete concepts* [5]. *Conceptual consistency* is a measure of how well LLMs understand the *relationships*

*between concepts*. Popular LLMs only have a *moderate amount* of conceptual consistency [6]. This suggests that these models may not have a deep understanding of the concepts they are able to answer questions about.

LLMs can be used for *concept extraction*, which is the process of identifying and extracting concepts from *text*. There are several ways to use LLMs for concept extraction. One way is to use a technique called *"named entity recognition" (NER)*. However, the performance of LLMs on NER is significantly below supervised baselines. This is due to the *gap between the two tasks, the NER and LLMs* [7]: the former is a sequence labeling task in nature while the latter is a text-generation model. Furthermore, LLMs often perform sub-optimal in *non-standard domains* [8], like the clinical domain, where a large gap between pre-training documents and target documents is observed.

LLMs can be used to extract *concept graphs* from text by first identifying the *concepts* in a text. Once the concepts have been identified, the *relationships* between them can be extracted and the concept graph constructed. In [9] an exhaustive quantitative and qualitative evaluation of LLMs for *concept graph construction* is performed, which consists of several tasks including *named entity recognition (NER)*, *relation extraction (RE)*, *event extraction (EE)*, and *entity linking (EL)*. The findings suggest that *GPT-4* outperforms ChatGPT in the majority of tasks and even surpasses fine-tuned models in certain reasoning and question-answering tasks.

LLMs are trained on massive datasets of text, and they can learn the meaning of *words, phrases, and even entire concepts*. This makes them a powerful tool for *concept learning*, which is the process of acquiring knowledge about a concept. There are a number of ways that LLMs can be used for concept learning. One way is to use them to *generate examples of the concept* being learned. Another way is to use them to *generate a probability distribution over the possible meanings of the concept*.

Human knowledge consists of both *symbolic (conceptual) knowledge* and *embodied (sensory) knowledge*. LLMs, however, are trained with natural-language text and can represent only *symbolic (conceptual) knowledge*. *Multimodal LLMs* [11-12], on the other hand, can process and generate text, images, and other types of data. They are capable of representing the *full range (conceptual and sensory)* of human knowledge. *Visual-language LLMs* are the most important multimodal LLMs. We discuss *conceptual understanding* in visual-language LLMs as well as major use of them for CODL including *concept extraction from image*, *concept graph extraction from image*, and *concept learning*.

# 2 Background

## 2.1 Concept-Oriented Deep Learning (CODL)

*Concept-oriented deep learning (CODL)* [1] is a machine learning approach that extends deep learning with *concept representations* and *conceptual understanding capability*. CODL addresses some of the major limitations of deep learning, such as interpretability, transferability, contextual adaptation, and requirement for lots of labeled training data. CODL is based on the idea that *concepts* are the foundation of human deep learning, understanding, and knowledge integration and transfer.

*CODL systems* are composed of three main components: concept graphs, concept representations, and concept representation learning systems. C*oncept graphs* are a knowledge base that contains information about concepts, such as their definitions, properties, and relationships to other concepts. *Concept representations* are low-dimensional vectors that represent the meaning of concepts. *Concept representation learning systems* learn concept representations from data, supporting incremental and continual learning.

Here are some of the *benefits* of using CODL systems:

- Interpretability: CODL systems can be more interpretable than traditional deep learning systems because they are based on *concepts*. Concepts are high-level representations of entities and their relationships, which can be easier for humans to understand than low-dimensional vectors.
- Transferability: CODL systems can be more transferable than traditional deep learning systems because they learn *concept representations* that are not specific to a particular task. This means that CODL systems can be used for a variety of tasks without having to be retrained from scratch.
- Contextual adaptation: CODL systems can be more adaptable to new tasks than traditional deep learning systems because they can use *concept graphs* to reason about the context of a new task. This allows CODL systems to generalize to new tasks more effectively.
- Less data requirement: CODL systems can require less labeled training data than traditional deep learning systems because they can learn concept representations from *unlabeled data* or *few exemplar data*. This makes CODL systems more scalable and cost-effective.

## 2.2 Dual Embodied-Symbolic Concept Representations (DESCR)

*Dual embodied-symbolic concept representations (DESCR)* [2] is an approach to concept representations that combines the strengths of embodied and symbolic representations. *Embodied representations* are grounded in sensory experience, while *symbolic representations* are language-based. The embodied level consists of *concept-oriented feature representations*, and the symbolic level consists of *concept graphs*. Embodied representations are modality specific; symbolic representations are amodal and language specific.

Here is how *DESCR representations* are formed:

1. Embodied representations: Embodied representations are learned from *sensory data*, such as images and videos. This data is processed by a neural network to create a representation of the object or scene in the data.
2. Symbolic representations: Symbolic representations are learned from *language text*. This data is processed by a language model to create a representation of the meaning of the text.
3. Fusion: The embodied and symbolic representations are fused together to create a DESCR representation. This representation captures both the *sensory and conceptual aspects* of the object or scene.

DESCR representations have several *advantages* over traditional deep learning representations:

- Interpretability: DESCR representations are more interpretable than traditional deep learning representations because they are similarly grounded in *sensory experience* but additionally formed from *conceptual (language) understanding*. This makes it easier for humans to understand how DESCR representations work and how they can be used to solve problems.
- Transferability: DESCR representations are more transferable than traditional deep learning representations because they are learned from both *sensory data* and *conceptual information*. This means that DESCR representations can be used for a variety of tasks without having to be retrained from scratch.
- Accuracy: DESCR representations are more accurate than traditional deep learning representations for a variety of tasks. This is because DESCR representations capture the *full range (sensorial and conceptual) of human knowledge*, which allows them to better understand the world.

## 2.3 Large Language Models (LLMs)

A *large language model (LLM)* [3-4] is a *deep learning model* with many parameters (typically billions of weights or more), trained on large quantities of *unlabeled text* using self-supervised learning or semi-supervised learning. LLMs are *general purpose models* which excel at a wide range of tasks, as opposed to being trained for one specific task (such as named entity recognition, sentiment analysis, or text classification).

Some of the most common *natural-language tasks and applications* that LLMs can perform include:

- Text translation: LLMs can be used to translate text from one language to another.
- Text summarization: LLMs can be used to summarize text.
- Text generation: LLMs can be used to generate text, such as news articles or creative writing.
- Question answering: LLMs can be used to answer questions about text.
- AI chatbots: LLMs can be used to create AI chatbots that can engage in natural conversations with humans.

LLMs are capable of performing these tasks, based on their internal knowledge stored in parameters during pre-training. However, LLMs do not promise *concept understanding* nor guarantee *conceptual consistency*, which could lead LLMs to generate factually wrong results. This is discussed in the next section.

## 3 Concept Understanding and Conceptual Consistency in LLMs

One of the key challenges in developing LLMs is *concept understanding*. LLMs need to be able to understand the meaning of words and phrases in order to generate accurate and meaningful text. However, this can be difficult, as many words and phrases have multiple meanings. For example, the word "Java" can refer to a brand of coffee, a programming language, or an island.

LLMs typically use a variety of *deep learning techniques* to understand concepts. One common technique is to use *word embeddings*. Word embeddings are vector representations of words that capture their meaning. For example, the word "Java" might have a word embedding that is similar to the word "Python". This allows LLMs to understand the relationship between words, even if they have different meanings. Another common technique is to use *supervised or unsupervised learning*, which can learn the relationship between words and concepts. This allows LLMs to understand the meaning of words and phrases in context.

Here are some examples of how concept understanding can *improve LLM tasks*:

- Text translation: By understanding the meaning of words and phrases in both languages, LLMs can generate more accurate translations.
- Text generation: By understanding the meaning of words and phrases, LLMs can generate new content that is both original and meaningful.
- Question answering: By understanding the meaning of words and phrases, LLMs can provide accurate and comprehensive answers to questions.

In [5] LLMs' ability to understand concepts, especially *abstract* and *concrete concepts*, is explored. It constructs a WordNet-based dataset containing a subset for abstract concepts and a subset for concrete concepts. It selects six LLMs and conducts a classic NLP task, *hypernym discovery*, as evidence of LLMs' comprehension ability in understanding concepts. The experimental results suggest that the LLM's understanding of *abstract concepts* is significantly weaker than that of concrete concepts. This is likely because abstract concepts are more difficult to define and are often based on subjective experiences. For example, the concept of "love" is difficult to define and can mean different things to different people.

### Conceptual Consistency

*Conceptual consistency* is a measure of how well LLMs understand the *relationships between concepts*. It is calculated by measuring the consistency of the LLM's responses to queries about conceptually related concepts. For example, if an LLM is asked "What is a mountain?" and it responds "A mountain is a large mammal," then its response is not conceptually consistent. Conceptual consistency is an important measure of LLM performance because LLMs with high conceptual consistency are more likely to produce accurate and meaningful responses to questions.

There are a number of ways to *improve* the conceptual consistency of LLMs. One way is to train the LLM on a dataset that includes information about the relationships between concepts. Another way is to use a technique called "prompt engineering" to guide the LLM's responses to questions. Prompt engineering involves carefully crafting the questions that are asked to the LLM in order to elicit the desired responses.

Here are some examples of how conceptual consistency can be used to *improve the performance of LLMs*:

- Text translation: By understanding the relationships between concepts in the source and target languages, LLMs can generate more accurate translations.
- Text generation: By understanding the relationships between concepts, LLMs can generate new content that is both creative and meaningful.
- Question answering: By understanding the relationships between concepts, LLMs can provide answers that are both accurate and comprehensive.

To compute conceptual consistency, [6] first extracts *background knowledge* for a question by traversing paths between concepts in a knowledge base. It then uses this background knowledge to predict an LLM's response to the anchor query. The *conceptual consistency* of the model is calculated as the accuracy of these predictions. It finds that conceptual consistency increases with the scale of the LLM used. However, popular LLMs only have a *moderate amount* of conceptual consistency. This suggests that these models may not have a deep understanding of the concepts they are able to answer questions about.

## 3.1 AI Chatbot Example

This example illustrates the critical importance of *concept understanding* and *conceptual consistency* to AI chatbots based on LLMs. For the example, we provide the same prompt to two new and prominent AI chatbots, but only one returns a satisfactory response. The prompt is:

*"People drink Java at the Java conference in Java. What is the first Java, the second, and the third?"*

The first chatbot, Bing Chat, returns the following response:

*"I'm not sure what you mean by the first Java, second Java and third Java. Could you please clarify your question?"*

Bing Chat doesn't understand the natural-language concept of "coreference" and fails to return a satisfactory response.

The second chatbot, Google Bard, returns the following satisfactory and intelligent response:

*"The first Java is the coffee, the second Java is the programming language, and the third Java is the island in Indonesia. The coffee is named after the island, and the programming language is named after the coffee. The conference is about the programming language, and*

*it is being held on the island. So, people are drinking coffee at the conference about the programming language that is named after the coffee that is named after the island."*

Google Bard understands the natural-language concept of "co-reference" as well as the domain concepts of "coffee, "programming language" and "island", with "Java" being an instance of each. It forms the response based on conceptual consistency: "drink Java" with "Java coffee", "Java conference" with "Java programming language", and "in Java" with "Java island".

## 4 Concept Extraction from Text with LLMs

LLMs can be used for *concept extraction*, which is the process of identifying and extracting concepts from text. LLMs are trained on large datasets of text, which allows them to learn the *relationships between words and phrases*. This knowledge can be used to identify concepts in new text, even if the text is not part of the dataset that the LLM was trained on.

There are several ways to use LLMs for concept extraction. One way is to use a technique called *"named entity recognition" (NER)*. NER is a process of identifying named entities in text, such as people, places, and events. LLMs can be used to improve the accuracy of NER by providing additional context about the text. Another way is to use a technique called *"relation extraction" (RE)*. RE is a process of identifying relationships between entities in text. LLMs can be used to improve the accuracy of RE, again, by providing additional context about the text. LLMs can also be used for concept extraction in a more creative way. For example, an LLM could be used to generate a list of *possible concepts* for a given piece of text. This could be useful for tasks such as brainstorming and research.

In the case of the *NER technique*, the performance of LLMs on NER is significantly below supervised baselines. This is due to the *gap between the two tasks, the NER and LLMs*: the former is a sequence labeling task in nature while the latter is a text-generation model. *GPT-NER* [7] bridges the gap by transforming the NER sequence labeling task to a generation task that can be easily adapted by LLMs. However, LLMs have the *hallucination issue*: they have a strong tendency to label NULL inputs as entities. To efficiently address this issue, GPT-NER uses a *self-verification strategy* by prompting LLMs to ask itself whether the extracted entities belong to a labeled entity tag. *Experiments* on five widely adopted NER datasets are conducted, and GPT-NER achieves comparable performances to fully supervised baselines. More importantly, GPT-NER exhibits a greater ability in the low-resource and few-shot setups. When the amount of training data is extremely scarce, GPT-NER performs significantly better than supervised models.

LLMs often perform sub-optimal in *non-standard domains*, like the clinical domain, where a large gap between pre-training documents and target documents is observed. The *CLIN-X (Clinical XLM-R)* LLM [8], using *(clinical) domain-adaptive pre-training*, outperforms other LLMs by a large margin for ten *clinical concept extraction* tasks from two languages. It highlights the importance of *specialized LLMs*, such as CLIN-X, for concept extraction in non-standard domains.

Here are some examples of how *specialized LLMs* can be used for concept extraction:

- A financial LLM could be used to identify *financial concepts* in news articles. This could help investors to make better financial decisions.
- A legal LLM could be used to identify *legal concepts* in legal documents. This could help lawyers to prepare for trials and other legal proceedings.
- A medical LLM could be used to identify *medical concepts* in patient records. This could help doctors to diagnose and treat patients more effectively.

## 4.1 AI Chatbot Example

In the example discussed in Section 3.1 AI Chatbot Example, the following prompt is provided:

*"People drink Java at the Java conference in Java. What is the first Java, the second, and the third?"*

In order for Google Bard to return the intelligent response shown there, it implicitly extracts from the prompt the following concepts (instances):

- "coreference" ("first Java" <-> "(drink) Java", "second (Java)" <-> "Java (conference)", "third (Java)" <-> "(in) Java")
- "People"
- "drink", "coffee ("Java")
- "at", "programming language" ("Java"), "conference"
- "in", "island" ("Java")

# 5 Concept Graph Extraction from Text with LLMs

LLMs can be used to extract *concept graphs* from text by first identifying the *concepts* in a text. This can be done using a variety of techniques, such as NER discussed in the previous section. Once the concepts have been identified, the *relationships* between them can be extracted using a variety of techniques, such as *dependency parsing* and *coreference resolution*.

The use of LLMs for concept graph extraction has a number of *advantages*. First, LLMs can be trained on large amounts of text data, which allows them to learn to identify and represent a wide range of concepts and relationships. Second, LLMs can be used to extract concept graphs from text that is not well-structured, such as free text or social media posts. Furthermore, LLMs can be used to extract concept graphs from text in a variety of languages.

Concept graphs can be used for a variety of *LLM applications*, such as:

- Natural language processing: Concept graphs can be used to improve the performance of natural language processing (NLP) tasks, such as *text translation, text summarization*, and *sentiment analysis*. For example, an LLM could be used to extract a concept graph from a document. The concept graph could then be used to improve the performance of text translation that is translating the document into another language.
- Question answering: Concept graphs can be used to answer questions about text. For example, an LLM could be used to extract a concept graph from a book. The concept graph could then be used to answer questions about the book, such as "What are the main characters' relations?"
- Knowledge discovery: Concept graphs can be used to discover new knowledge from text. For example, an LLM could be used to extract a concept graph from a collection of scientific papers. The concept graph could then be used to identify new relationships between concepts, which could lead to new discoveries.

In [9] an exhaustive quantitative and qualitative evaluation of LLMs for *Knowledge Graph (KG), i.e., concept graphs*, construction and reasoning is performed. Typically, KG construction consists of several tasks, including *named entity recognition (NER)*, r*elation extraction (RE)*, e*vent extraction (EE)*, and e*ntity linking (EL)*; KG reasoning refers to l*ink prediction (LP)*. The work investigates the potential applicability of LLMs, exemplified by ChatGPT and GPT-4, in KG construction and KG reasoning tasks. Specifically, it investigate the *zero-shot* and *one-shot* performance of LLMs on entity, relation and event extraction, link prediction, and question answering to assess LLMs' potential applications within the KG domain. The findings suggest that *GPT-4* outperforms ChatGPT in the majority of tasks and even surpasses fine-tuned models in certain reasoning and question-answering tasks.

## 5.1 AI Chatbot Example

In Google Bard, concept graphs can be extracted in the form of "concept graph", "knowledge graph" or "ontology". A *concept graph* consists of concepts (nodes, with attributes) and relations (edges). A *knowledge graph* consists of entities (nodes, with attributes) and relationships (edges), which represent knowledge (factual and conceptual) about the world. An *ontology* consists of categories, entities, relations and axioms, which specify the formal conceptualization of a domain.

The example utilizes *concept-oriented prompting* which focuses on *concept understanding*.

### Concept Graph

Concept graph extraction starts with concept identification. The initial prompt used is:

*People drink Java at the Java conference in Java. Please identify the concepts involved.*

Bard responds with:

The concepts involved in the statement "People drink Java at the Java conference in Java" are:

- **Metaphor:** The word "Java" is used to refer to both the programming language and the coffee drink. This is an example of a metaphor, a figure of speech in which one thing is likened to another.
- **Polysemy:** The word "Java" has multiple meanings. In this case, it is used to refer to both the programming language and the coffee drink. This is an example of polysemy, a linguistic phenomenon in which a word has multiple meanings.
- **Wordplay:** The statement is a play on words, because it uses the multiple meanings of the word "Java" to create a humorous effect.

Since our focus is on world concepts, not linguistic concepts, we revise the prompt to:

*People drink Java at the Java conference in Java. Please identify the world concepts involved.*

Bard then responds:

Sure, here are the world concepts involved in the sentence "People drink Java at the Java conference in Java":

- **Java:** This is the name of the programming language that the conference is about. It is also the name of a coffee drink.

- **Conference:** This is a meeting of people who share a common interest. In this case, the common interest is Java.
- **Drink:** This is the action of consuming a liquid. In this case, the liquid is Java coffee.
- **Place:** This is the location of the conference. In this case, the conference is in Java, which is a coffee-growing region in Indonesia.

The response, however, misses the fact that Java is also an island. We augment the prompt with the missing fact:

*Java is also an island. Please update the world concepts involved in the original sentence.*

Bard responds by adding the missing fact:

Sure, here are the world concepts involved in the sentence "People drink Java at the Java conference in Java" after we take into account that Java is also an island:

- **Java:** This is the name of the programming language that the conference is about. It is also the name of a coffee drink, and it is also the name of an island in Indonesia.
- **Conference:** This is a meeting of people who share a common interest. In this case, the common interest is Java.
- **Drink:** This is the action of consuming a liquid. In this case, the liquid could be Java coffee or Javanese coffee, which is a type of coffee grown on the island of Java.
- **Place:** This is the location of the conference. In this case, the conference could be held on the island of Java, or it could be held in a coffee shop in Java.

"People" is missing from the list of concepts. We augment the prompt with the missing concept:

*You missed people. Please update the world concepts involved in the original sentence.*

Bard responds by adding the missing concept:

You're right, I did miss people. Here are the world concepts involved in the sentence "People drink Java at the Java conference in Java", including people:

- **People:** This is the group of humans who are attending the conference.
- **Java:** This is the name of the programming language that the conference is about. It is also the name of a coffee drink, and it is also the name of an island in Indonesia.
- **Conference:** This is a meeting of people who share a common interest. In this case, the common interest is Java.
- **Drink:** This is the action of consuming a liquid. In this case, the liquid could be Java coffee or Javanese coffee, which is a type of coffee grown on the island of Java.
- **Place:** This is the location of the conference. In this case, the conference could be held on the island of Java, or it could be held in a coffee shop in Java.

Now, we are ready to generate the concept graph with the prompt:

*Given the concepts identified, please generate the concept graph for the original sentence.*

Here is Bard's response:

> Sure, here is the concept graph for the original sentence "People drink Java at the Java conference in Java":
>
> ```
> People --> drink --> Java
> Java --> programming language
> Java --> coffee drink
> Java --> island in Indonesia
> Conference --> meeting of people
> Conference --> about Java
> Place --> location of conference
> Place --> island of Java
> ```

## Knowledge Graph

The initial prompt used for knowledge graph extraction is:

*People drink Java at the Java conference in Java. Please generate the knowledge graph.*

Bard returns the following response:

> Sure, here is the knowledge graph:
>
> * **Entity:** Java conference
>   * **Attributes:**
>     * Location: Java
>     * Topics: Java
>     * Participants: Java developers
>   * **Relationships:**
>     * Has attendees: People
>     * Serves beverages: Java
> * **Entity:** People
>   * **Attributes:**
>     * Drink: Java
>   * **Relationships:**
>     * Attends Java conferences
> * **Entity:** Java
>   * **Attributes:**
>     * A programming language
>     * A beverage
>   * **Relationships:**
>     * Is the topic of Java conferences
>     * Is drunk by people at Java conferences

Since the response misses the fact that Java is also an island, we augment the prompt with the missing fact:

*Java is also an island. Please regenerate the knowledge graph.*

Bard revises the response by adding the missing fact:

Sure, here is the updated knowledge graph:

* **Entity:** Java conference
    * **Attributes:**
        * Location: Java island
        * Topics: Java
        * Participants: Java developers
    * **Relationships:**
        * Has attendees: People
        * Serves beverages: Java (beverage)
* **Entity:** People
    * **Attributes:**
        * Drink: Java (beverage)
    * **Relationships:**
        * Attends Java conferences
* **Entity:** Java
    * **Attributes:**
        * A programming language
        * A beverage
        * An island
    * **Relationships:**
        * Is the topic of Java conferences
        * Is drunk by people at Java conferences
        * Is the location of Java conferences

## Ontology

The prompt used for ontology extraction is:

*People drink Java at the Java conference in Java. Please generate the ontology.*

Bard returns the following response:

Sure, here is the ontology for the statement "People drink Java at the Java conference in Java":

Ontology: Java Conference Ontology

Classes:

* Person
* Java Conference
* Java

Object Properties:

* drink
* attend
* located in

Data Properties:

* name

* date
* location

Individuals:

* John Doe: a Person
* JavaOne: a Java Conference
* Java programming language: a Java

Facts:

* John Doe drinks Java at JavaOne.
* JavaOne is located in Java.

This ontology captures the following information about the statement:

- People are individuals of the class Person.
- Java conferences are individuals of the class Java Conference.
- Java is a programming language and an individual of the class Java.
- The act of drinking Java is an instance of the object property drink.
- The act of attending a Java conference is an instance of the object property attend.
- The location of a Java conference is an instance of the object property located in.

The name, date, and location of a Java conference are data properties.

## 6 LLMs for Concept Learning

LLMs are trained on massive datasets of text, and they can learn the meaning of *words, phrases, and even entire concepts.* This makes them a powerful tool for *concept learning*, which is the process of acquiring knowledge about a concept.

There are a number of ways that LLMs can be used for concept learning. One way is to use them to *generate examples of the concept* being learned. For example, if you are trying to learn the concept of "dog," you could use an LLM to generate a list of sentences that contain the word "dog." Another way is to use them to *generate a probability distribution over the possible meanings of the concept*. For example, if you are trying to learn the concept of "love," you could use an LLM to generate a probability distribution over the possible meanings of the word "love."

LLMs have a number of *advantages* for concept learning. First, they can be used to learn the meaning of concepts that are difficult or impossible to define in a traditional way. Second, they can be used to learn the meaning of concepts from text that is not explicitly labeled with the meaning of the concept. Third, they can be used to learn the meaning of concepts from text that is noisy or incomplete. Lastly, they can be used to learn the meaning of concepts in multiple languages and translate between them.

As *examples* for concept learning, LLMs can be used to:

- Identify and classify concepts. LLMs can be trained on a dataset of text that includes examples of different concepts. They can then be used to identify and classify new instances of those concepts.
- Learn the relationships between concepts. LLMs can be trained on a dataset of text that includes examples of how different concepts are related to each other. They can then be used to learn the relationships between new instances of those concepts.
- Generate new concepts. LLMs can be used to generate new concepts by combining existing concepts in new ways. This can be useful for tasks such as brainstorming, research, or generating creative text.

In [10] an *LLM* is used to model *learning of abstract symbolic concepts* by performing *Bayesian inference* over utterances in natural language. The work uses an LLM as a proposal distribution, fits a prior to *human data* to better model human learners, and evaluates on both *generative and logical concepts*. The *symbolic concept learning* model expresses its concepts in *natural language*, even when the learning problem does not involve natural language, for two reasons. First, language is an effective representation for many human concepts. It is compositional, richly expressive, and regularizes the learner toward natural generalizations. Second, LLMs can be used to efficiently infer natural language concepts.

## 7 Multimodal LLMs for Multimodal (Dual Symbolic-Embodied) Concepts

Human knowledge consists of both *symbolic (conceptual) knowledge* and *embodied (sensory) knowledge*, as discussed in Section 2.2 Dual Embodied-Symbolic Concept Representations (DESCR). *LLMs*, however, are trained with natural-language text and can represent only *symbolic (conceptual) knowledge*. *Multimodal LLMs* [11-12], on the other hand, can process and generate text, images, and other types of data. They are trained on massive datasets of multimodal data, which allows them to learn the relationships between different modalities. This makes them capable of tasks that would be impossible for text-only LLMs, such as describing images and generating captions for videos. Multimodal LLMs, therefore, are capable of representing the *full range (conceptual and sensory)* of human knowledge.

Here are some of the *benefits* of using multimodal LLMs:

- Enhanced user experience: Multimodal LLMs can create more realistic and engaging user experiences by incorporating *sensory data* into their outputs.

- Increased accuracy: Multimodal LLMs can learn the relationships between *different modalities*, which allow them to make more accurate predictions and generate more realistic outputs.
- Increased creativity: Multimodal LLMs can generate new ideas and concepts by combining information from *different modalities*.
- Improved performance: Multimodal LLMs can perform tasks that would be impossible for text-only LLMs, such as describing *images*, and generating captions for *videos*.

## 7.1 Visual-Language LLMs

*Visual-language LLMs* are the most important multimodal LLMs. In recent years, there has been a growing interest in using LLMs for visual-language tasks. These tasks involve understanding the *relationship between text and images*, and using this understanding to perform *tasks* such as:

- Image captioning: LLMs can be used to generate captions for images. This can be useful for people who want to quickly understand the content of an image.
- Text-to-image synthesis: LLMs can be used to synthesize images from text descriptions. This can be used for creative applications, such as generating art.
- Visual question answering: LLMs can be used to answer questions about images. This can be useful for people who want to learn more about an image.

There are a number of *advantages* to using LLMs for visual-language tasks. First, LLMs have been trained on massive datasets of text, which gives them a deep understanding of both languages and the world. Second, LLMs are able to learn long-range dependencies between words and concepts, which is essential for understanding the relationship between text and images. Third, LLMs are able to generate creative and informative text, which is useful for tasks such as image captioning.

### Conceptual Understanding in Visual-Language LLMs

Visual-language LLMs have achieved great success in a variety of downstream tasks, such as image captioning, image question answering, and visual dialogue. However, it is not clear if these models have *conceptual understanding* of the content they are processing. In [13] a novel framework for probing and improving conceptual understanding of visual-language LLMs is proposed. The work introduces a novel benchmark dataset for probing *three aspects of conceptual understanding of an image*:

- Relational understanding: The ability to understand the relationships between entities in an image.
- Compositional understanding: The ability to understand how entities in an image can be combined to form new concepts.
- Contextual understanding: The ability to understand how the context of an image can affect the interpretation of its content.

It finds that visual-language LLMs are able to achieve good performance on tasks that require *relational understanding*, such as image question answering. However, they are less successful on tasks that require *compositional and contextual understanding*, such as visual question generation. This suggests that visual-language LLMs may not have a deep understanding of the content they are processing.

## Concept Extraction from Text and Image with Visual-Language LLMs

*Concept extraction* is the process of identifying and extracting concepts from text or image. This is a challenging task, as concepts can be represented in a variety of ways, both in text and in image. *Concept extraction from text* has been discussed in Section 4 Concept Extraction from Text with LLMs.

Visual-language LLMs can be used for *concept extraction from image* in a number of ways. One way is to use the LLM to *generate a natural language description* of an image. This description can then be analyzed to identify the concepts that are present in the image. Another way is to use the LLM to *answer questions* about an image. The questions that are asked can be designed to elicit information about specific concepts. For example, a question like "What is the object in the foreground?" can be used to extract the concept of "object" from the image. Finally, both ways can be combined by using the LLM to generate a natural language description of an image, and then using the LLM to answer questions about the image. The combination of the natural language description and the answers to the questions can then be used to identify the concepts that are present in the image.

Here are some examples of how concept extraction from image can be used in *real-world applications*:

- Image search: Concept extraction can be used to improve the accuracy of image search. By identifying the concepts that are present in an image, visual-language LLMs can help to match the image to relevant search results.

- Virtual assistants: Concept extraction can be used to improve the capabilities of virtual assistants. By understanding the concepts that are present in a user's query with image, virtual assistants can provide more relevant and informative responses.

## Concept Graph Extraction from Text and Image with Visual-Language LLMs

*Concept graph extraction* is the task of extracting a graph of concepts from text or image. Concept graph extraction from text has been discussed in Section 5 Concept Graph Extraction from Text with LLMs.

There are a number of different approaches to *concept graph extraction from image*, including:

- Text-based approaches: These approaches use natural language processing techniques to extract concepts from the *text associated with an image*.
- Image-based approaches: These approaches use computer vision techniques to extract concepts from the *image itself*.
- Hybrid approaches: These approaches combine text-based and image-based approaches to extract concepts from both the image and the associated text.

Concept graph extraction from image can be used for a variety of *tasks*, such as:

- Image understanding: Concept graphs can be used to represent the conceptual structure of an image, which can then be used to understand the meaning of the image.
- Visual question answering: Concept graphs can be used to represent the conceptual structure of a question involving image, which can then be used to answer the question.
- Visual dialogue: Concept graphs can be used to represent the conceptual structure of a dialogue involving image, which can then be used to generate more natural and engaging dialogue.

## Visual-Language LLMs for Concept Learning

Visual-language LLMs can be used for *concept learning* in a number of ways, including:

- Learn about new concepts: Visual-language LLMs can be used to learn about new concepts by *generating examples* of the concepts. For example, a visual-language LLM could generate a set of images of dogs to learn about the concept of "dog".
- Explore the relationships between concepts: Visual-language LLMs can be used to explore the relationships between concepts by asking them questions about the concepts. For example, a visual-language LLM could be asked "What is the difference between a dog and a cat?" and could generate a set of contrasting images of dogs and cats to learn about the difference.
- Generate new concepts: Visual-language LLMs can be used to generate new concepts by combining existing concepts in new ways. For example, a visual-language LLM could be used to generate the concept of "a dog wearing a hat" by combining the concepts of "dog" and "hat" and generate example images of the new concept. This can be useful for tasks such as brainstorming, research, or generating creative text.

Here are some of the *benefits* of using visual-language LLMs for concept learning:

- They can learn from large and diverse datasets: Visual-language LLMs can be trained on large and diverse datasets of text and images, which allows them to learn about a wide range of concepts.
- They can learn about concepts in a multimodal way: Visual-language LLMs can learn about concepts by incorporating both text and images, which gives them a more complete understanding of the concepts.

# 8 Conclusion

With concept understanding and conceptual consistency, LLMs are excellent, though implicit, concept representation learning systems which can learn symbolic (conceptual) concept representations from text and support incremental and continual learning. Multimodal LLMs, furthermore, can (implicitly) learn multimodal (dual symbolic-embodied) concept representations and thus capture the full range (conceptual and sensorial) of human knowledge.

As such, LLMs are a promising new technology for CODL. They can be used for major CODL tasks including concept extraction from text, concept graph extraction from text, and concept learning. Visual-language LLMs (the most important multimodal LLMs), moreover, can be used for CODL including concept extraction from image, concept graph extraction from image, and concept learning. While uses of LLMs for CODL are valuable standalone, they are particularly valuable as part of LLM applications such as AI chatbots.

**Acknowledgement:** Thanks to my wife Hedy (郑期芳) for her support.